\documentclass[10pt,twocolumn,letterpaper]{article}
\pdfoutput=1
\usepackage{iccv}
\usepackage{times}
\usepackage{epsfig}
\usepackage{graphicx}
\usepackage{amsmath}
\usepackage{amssymb}
\usepackage{enumitem}
\usepackage[accsupp]{axessibility}  

\usepackage[pagebackref=true,breaklinks=true,letterpaper=true,colorlinks,bookmarks=true]{hyperref}

\iccvfinalcopy 

\def\iccvPaperID{11289} 

\ificcvfinal\fi

\begin{document}

\title{Towards High-Quality Specular Highlight Removal \\by Leveraging
Large-Scale Synthetic Data}


\author{Gang Fu$^1$, Qing Zhang$^2$, Lei Zhu$^{3,4}$, Chunxia Xiao$^5$, and
  Ping Li$^{1,}$\thanks{Corresponding author.}\\
  {}$^1$The Hong Kong Polytechnic University, Hong Kong \\
  {}$^2$Sun Yat-sen University, Guangzhou, China \\
  {}$^3$The Hong Kong University of Science and Technology
  (Guangzhou), Guangzhou, China\\
  {}$^4$The Hong Kong University of Science and Technology, Hong Kong \\
  {}$^5$School of Computer Science, Wuhan University, Wuhan, China\\
}

\maketitle
\ificcvfinal\thispagestyle{empty}\fi

\begin{abstract}

  This paper aims to remove specular highlights from a single
  object-level image. Although previous methods have made some
  progresses, their performance remains somewhat limited, particularly
  for real images with complex specular highlights. To this end, we
  propose a three-stage network to address them. Specifically, given
  an input image, we first decompose it into the albedo, shading, and
  specular residue components to estimate a coarse specular-free
  image. Then, we further refine the coarse result to alleviate its
  visual artifacts such as color distortion. Finally, we adjust the
  tone of the refined result to match that of the input as closely
  as possible. In addition, to facilitate network training and
  quantitative evaluation, we present a large-scale synthetic dataset
  of object-level images, covering diverse objects and illumination
  conditions. Extensive experiments illustrate that our network is
  able to generalize well to unseen real object-level images, and even
  produce good results for scene-level images with multiple background
  objects and complex lighting.

\end{abstract}

\section{Introduction}


Specular highlights are very common in the real world, but they are
usually undesirable in photographs, since they can degrade the image
quality. In daily life, users often want to achieve the specular-free
image from an image. For example, specular highlights in facial or
document images sweep away skin details or meaningful texture patterns
which are very important to users. Removing specular highlights from a
single image enables recovering visual content with better
perceptibility. Moreover, it has many related applications such as
recoloring \cite{beigpour-2011-objec}, light source estimation
\cite{jiddi-2020-detec-specul}, recognition of specular objects
\cite{netz-2012-recog-using}, and intrinsic image decomposition
\cite{zhang-2021-unsup-intrin}. Thus, specular highlight removal is a
long-standing and challenging problem in computer vision and computer
graphics.

\begin{figure*}[t]
  \centering
  \begin{tabular}{c}
    \makebox[0pt]{\includegraphics[width=0.99\textwidth]{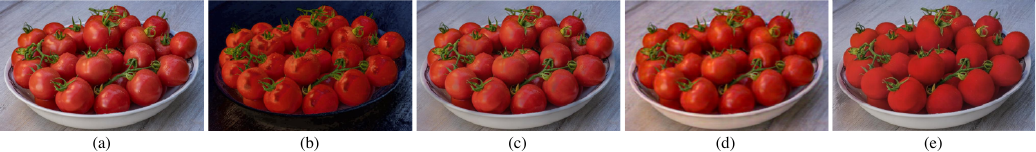}}
  \end{tabular}
  \caption{Visual comparison of our method against state-of-the-art
    methods on a challenging image with nearly white material
    surfaces. (a) Input. (b) Yang \textit{et al.}
    \cite{yang-2015-effic-robus}. (c) Fu \textit{et al.}
    \cite{fu-2021-multi-task}. (d) Wu \textit{et al.}
    \cite{wu-2021-singl-image-a}. (e) Ours.}
  \label{fig:teaser}
\end{figure*}

To address this problem, researchers have proposed various specular
highlight methods. They can be roughly divided into two categories:
traditional methods \cite{tan-2005-separ-reflec,
  yang-2015-effic-robus, kim-2013-specul-reflec,
  son-2020-towar-specul} based on intensity and chromaticity analysis
as well as optimization, and deep learning-based methods
\cite{yi-2019-lever-multi, fu-2021-multi-task, wu-2021-singl-image-a}.
However, the traditional methods often produce unsatisfactory or even
poor results with visual artifacts such as black color block and detail
missing; see Figure~\ref{fig:teaser}(b). The main reason is that they
fail to capture high-level semantic information to recover the missing
colors and details underneath specular highlights using those
meaningful and reliable information from the non-highlight region. In
addition, although the deep learning-based methods have achieved
certain performance improvement, they may still produce unsatisfactory
results with visual artifacts such as illumination residue and color
distortion; see Figure~\ref{fig:teaser}(c)(e). It is partly attributed
to the fact that they are trained on relatively simple images in which
materials and illumination conditions are not diverse enough, leading
to their limited generalization to unseen images.

We in this paper propose a three-stage specular highlight removal
network, consisting of (i) physics-based specular highlight removal,
(ii) specular-free refinement, and (iii) tone correction. In the first
stage, based on a physics-based image formation model, we decompose an
input image into its albedo, shading, and specular residue components,
and then estimate a coarse specular-free image. In the second stage,
we further refine the coarse result to alleviate visual artifacts for
improving the quality. In the third stage, we adjust the tone of the
refined result to produce the final result with the similar tone of
the input. In addition, to facilitate network training and
quantitative evaluation, we build a large-scale synthetic dataset
rendered by software using diverse 3D models and real HDR environment
maps. Figure~\ref{fig:teaser} presents the visual comparison on a real
image. As shown, our method is able to produce high-quality
specular-free images without noticeable artifacts encountered by
previous methods. Below, we summarize the major contributions of our
work.

\begin{itemize}[topsep=0pt]
\renewcommand\labelitemi{$\bullet$}
  \setlength{\itemsep}{0pt}
  \setlength{\parsep}{0pt}
  \setlength{\parskip}{4pt}
\item We propose a three-stage specular highlight removal network to
  progressively eliminate multiple types of visual artifacts such as
  color distortion and tone inconsistency.
\item We present a large-scale synthetic dataset of object-level
  images, in which each specular highlight image has corresponding
  ground truth albedo, shading, specular residue, and specular-free images.
\item We conduct extensive experiments on existing datasets and our new dataset,
  and demonstrate that our method achieves better
  quantitative and qualitative results than state-of-the-art methods.
\end{itemize}

\section{Related Work}
\label{sec:relatedwork}

\noindent\textbf{Single-Image methods}. Early methods are mostly
based on chromaticity propagation or optimization. Tan and Ikeuchi
\cite{tan-2005-separ-reflec} proposed to remove specular highlights
via iteratively comparing the intensity logarithmic differentiation of
an input image and its specular-free image. Yang \textit{et al.}
\cite{yang-2015-effic-robus} proposed to use the bilateral filter to
propagate information from the diffuse region to the specular
highlight region. Kim \textit{et al.} \cite{kim-2013-specul-reflec}
formulated specular highlight removal as a MAP optimization problem
based on the priors of specular highlights in the real world. However,
these methods may produce unsatisfactory results with visual artifacts
such as black color block, resulting in unrealistic appearances. To
alleviate the issue, Liu \textit{et al.} \cite{liu-2015-satur-preser}
proposed a two-step method in which an over-saturated specular-free
image is first produced by global chromaticity propagation, and then
recovered its saturation via an optimization framework. Guo \textit{et
  al.} \cite{guo-2018-singl-image} proposed a sparse and low-rank
reflection model for specular highlight removal. However, they may
fail to effectively recover the missing content underneath specular
highlights.

Subsequently, researchers have proposed various deep learning-based
methods. Shi \textit{et al.} \cite{shi-2017-learn-non} proposed a
unified framework that can simultaneously estimate the albedo,
shading, and specular residue components from a single object-level
image. However, it fails to generalize well to real images with
complex specular highlights. Yi \textit{et al.}
\cite{yi-2019-lever-multi} proposed to leverage multi-view image sets
(\textit{i.e.}, customer product photos) to perform specular highlight
removal in an unsupervised way. Fu \textit{et al.}
\cite{fu-2021-multi-task} proposed a multi-task network for joint
specular highlight detection and removal based on a region-aware
specular highlight image formation model. Wu \textit{et al.}
\cite{wu-2021-singl-image-a} proposed a GAN-based network for specular
highlight removal using specular highlight detection map as guidance.
Jin \textit{et al.} \cite{jin-2022-estim-reflec} proposed to estimate
the reflectance layer from a single image with shadows and specular
highlights. Although these methods achieve good results, their
performance is often limited, particularly for real images with adverse
factors such as achromatic material surfaces and complex illumination
conditions. In contrast, our three-stage method is able to effectively
address previous challenging images.

\vspace{0.5em} \noindent\textbf{Multi-Image and Normal-Based Methods}.
Researchers have proposed various multi-image and normal-based methods
to more robustly remove specular highlights. Guo \textit{et al.}
\cite{guo-2014-robus-separ} proposed to remove specular highlights for
superimposed multiple images. Wei \textit{et al.}
\cite{wei-2018-specul-highl} proposed a unified framework of specular
highlight removal and light source position estimation
by assuming that surface geometry is known.
Li \textit{et al.}
\cite{li-2017-specul-highl} proposed a method for specular highlight
removal in facial images that may contain varying illumination colors,
with the help of facial surface normals. Although these
methods can produce promising results, the requirement of multiple
images or extra auxiliary cues limits their applicability.

\vspace{0.5em}\noindent\textbf{Benckmark Datasets}. Grosse \textit{et
  al.} \cite{grosse-2009-groun} presented the MIT intrinsic images
dataset, including 20 object-level images and their corresponding
ground truth intrinsic images. However, these images are not
sufficient to support network training. Shi \textit{et al.}
\cite{shi-2017-learn-non} rendered a large-scale synthetic dataset for
non-Lambertian intrinsic image decomposition by software. Although
this dataset includes a large amount of images, many of them do not
have obvious and meaningful specular highlights for our task.
Recently, Fu \textit{et al.} \cite{fu-2021-multi-task} presented a
real dataset simultaneously for specular highlight detection and
removal, produced by a series of image processing algorithms on the
multi-illumination dataset IIW \cite{murmann-2019-datas-multi}. At the
same time, Wu \textit{et al.} \cite{wu-2021-singl-image-a} also built
a real paired specular-diffuse image dataset via the
cross-polarization photography technique. However, objects and
illumination conditions in these two datasets are somewhat limited for
network training, leading to the unsatisfactory generalization to
unseen images. In contrast, we present a large-scale synthetic dataset
of object-level images, which covers diverse objects and illumination
conditions, and thus contains various appearances of specular
highlights.

\section{Methodology}
\label{sec:method}

\subsection{Overview}
\label{subs:overview}

Figure~\ref{fig:framework} presents the pipeline of our three-stage
framework. It consists of three stages: (i) physics-based specular
highlight removal; (ii) specular-free refinement; and (iii) tone
correction. Specifically, in the first stage (see (a)), we decompose
an input image into its albedo and shading using two encoder-decoder
networks ($E_a$-$D_a$ for albedo, and $E_s$-$D_s$ for shading). Then,
the specular-free image can be estimated by multiplying the albedo and
shading. In the second stage (see (b)), we feed the coarse result
along with the input into an encoder-decoder network
($E_r$-$D_r$) to further refine it to alleviate visual artifacts. In
the third stage (see (c)), we feed the refined result along with the
input and its specular residue image into an encoder-decoder network
($E_c$-$D_c$) to adjust its tone so that it has the similar tone as
the input as much as possible. Figure~\ref{fig:ablationstudy} validate
the effectiveness of each stage in our framework.

\subsection{Physics-Based Specular Highlight Removal}
\label{subs:physicshighlightremoval}

\begin{figure}[t]
  \centering
  \begin{tabular}{c}
    \makebox[0pt]{\includegraphics[width=0.99\columnwidth]{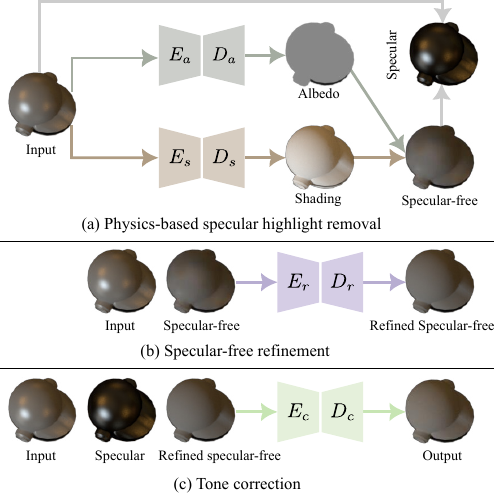}}
  \end{tabular}
  \caption{The pipeline of our three-stage specular highlight removal framework.}
  \label{fig:framework}
\end{figure}

According to the dichromatic reflection model \cite{shafer-1985}, an
input image $I$ can be decomposed into its intrinsic images
\footnote{Throughout the paper we use the terms \textit{albedo} and
  \textit{shading} loosely for simplicity. Actually, \textit{albedo}
  and \textit{shading} refer to diffuse albedo and diffuse shading,
  respectively.}, expressed as
\begin{equation}
I=A\times S + R\,,
  \label{eq:dichromaticmodel}
\end{equation}
where $A$, $S$, and $R$ are albedo, shading, and specular residue, respectively.
Based on the physical image formation model in
Eq.~(\ref{eq:dichromaticmodel}), we propose the Physics-based Specular
Highlight Removal stage (PSHR) to recover the intrinsic images from an
input image. Figure~\ref{fig:framework}(a) illustrates the mechanism
of PSHR. Specifically, given an input image, we use an encoder-decoder
network ($E_a$-$D_a$) to estimate albedo, and another one ($E_s$-$D_s$) to estimate
shading. The specular-free (\textit{i.e.}, diffuse) image $D$ is
estimated by
\begin{equation}
D=A \times S\,,
  \label{eq:diffuseestimation}
\end{equation}
With Eqs.~(\ref{eq:dichromaticmodel}) and
(\ref{eq:diffuseestimation}), we can yield the specular residue $R$ by
\begin{equation}
R=I-D\,.
  \label{eq:residueestimation}
\end{equation}
To facilitate the network training of PSHR, we present a large-scale
synthetic dataset of object-level images for specular highlight
removal (named SSHR). Now, we detail it.

\vspace{0.5em} \noindent\textbf{Dataset}. To the best of our
knowledge, SHIQ \cite{fu-2021-multi-task} and PSD
\cite{wu-2021-singl-image-a} are only two publicly available real
datasets for specular highlight removal. However, they suffer from the
following three issues. First, the quantity of objects is quite small,
and the images were captured in controllable laboratory environments
with limited illumination conditions. Second, a pair of specular
highlight and specular-free images may not be aligned well, since the
camera shakes caused by itself or hand touch during the process of
capturing data. Third, even a well-aligned pair of specular highlight
and specular-free images may have inconsistent color and shading,
since the environmental lighting may have a subtle fluctuation over
time and the camera exposure may vary. In addition, Shi \textit{et
  al.} \cite{shi-2017-learn-non} presented a large-scale synthetic
dataset for non-Lambertian intrinsic image decomposition. However,
most input images in it are not with obvious and meaningful specular
highlights, and thus are not well-suited for our task. Note that this
dataset is currently not publicly available.

\begin{figure}[t]
  \centering
  \begin{tabular}{c}
    \makebox[0pt]{\includegraphics[width=0.99\columnwidth]{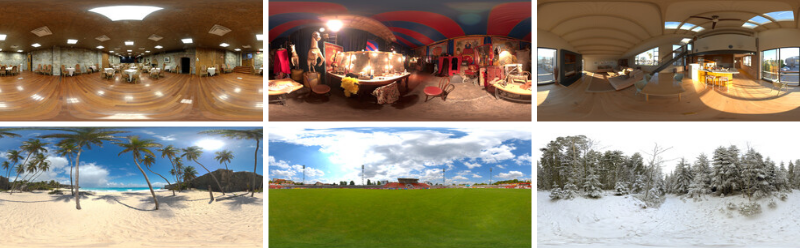}}
  \end{tabular}
  \caption{Example environment maps for our rendering. Top:
    indoor scenes. Bottom: outdoor scenes.
    }
  \label{fig:envshow}
\end{figure}

\begin{figure}[t]
  \centering
  \begin{tabular}{c}
    \makebox[0pt]{\includegraphics[width=0.99\columnwidth]{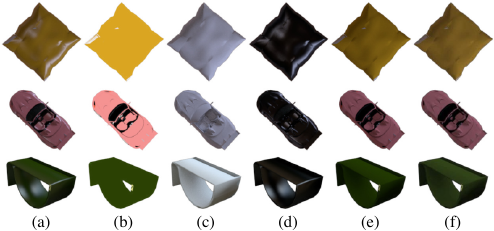}}
  \end{tabular}
  \caption{Example image groups in our dataset. (a) Input. (b) Albedo.
    (c) Shading. (d) Specular residue. (e) Ground truth. (f) Tone
    correction version of (e).}
  \label{fig:datashow}
\end{figure}

To this end, we built a large-scale synthetic dataset tailored for
specular highlight removal. Specifically, to render the data, we first
picked up 1500 3D models with their albedo texture maps from several
common categories (such as {car}, {bus}, {container}, and {sofa}) of
the large-scale 3D shape dataset ShapeNet \cite{chang-2015-shapen}.
Then, we collected 90 HDR environment maps from the Internet
\footnote{http://www.hdrlabs.com/sibl/archive.html.}, which includes
indoor and outdoor scenes with diverse material surfaces and
illumination conditions. Figure~\ref{fig:envshow} presents example
environment maps. Finally, we used an open-source render software
Mitsuba \cite{jakob-2010-mitsub} and adopted the modified Phong
reflection model \cite{phong-1975-illum-comput} to render object
models with various environment maps to generate photo-realistic
shading and specular residue appearance. According to the rendered
results, the specular-free and input images can be obtained via
Eqs.~(\ref{eq:diffuseestimation}) and (\ref{eq:dichromaticmodel}),
respectively. Finally, we randomly split the collected 1500 models
into $1300$ models for training and $200$ for testing. In total, we
have 117,000 training images and 18,000 testing images.
Figure~\ref{fig:datashow} shows example image groups in our dataset.

\vspace{0.5em}\noindent\textbf{Loss Function}. The total loss for
physics-based specular highlight removal $\mathcal{L}^{\text{PSHR}}$
is defined as
\begin{equation}
\mathcal{L}^{\text{PSHR}}=||A-\hat{A}||^2+||S-\hat{S}||^2+||I-D_1 - \hat{R}||^2\,,
  \label{eq:IDloss}
\end{equation}
where $\hat{A}$, $\hat{S}$, and $\hat{R}$ are the ground truths of the
estimated albedo $A$, shading $S$, and specular residue $R$, respectively;
and $D_1=A\times S$ is the estimated specular-free image. The
rightmost term of Eq.~(\ref{eq:IDloss}) is to encourage the estimated
specular residue image (\textit{i.e.}, $I-D_1$) to be similar
with its ground truth as much as possible.

\subsection{Specular-Free Refinement}
\label{sec:rashr}

\begin{figure}[t]
  \centering
  \begin{tabular}{c}
    \makebox[0pt]{\includegraphics[width=0.99\columnwidth]{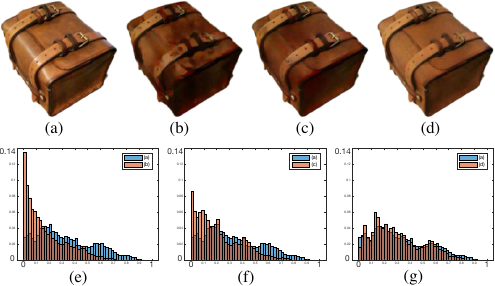}}
  \end{tabular}
  \caption{Ablation study that demonstrates the effectiveness of each
    stage in our framework. (a) Input. (b)-(d) Resulting specular-free
    images produced by the first, second, and third stages in our
    framework, respectively. (e)-(f) Histogram comparison between (a)
    and (b)-(d), respectively. Note that the abscissa and ordinate
    axes indicate the pixel intensity value and the ratio of the
    number of target pixels and the total number of pixels in an image.}
  \label{fig:ablationstudy}
\end{figure}

The first stage, PSHR, has two drawbacks. First, it tends to overly
remove specular highlights and produce visual artifacts such as color
distortion and black color block; see
Figure~\ref{fig:ablationstudy}(b). Second, the estimation of
specular-free image by Eq.~(\ref{eq:diffuseestimation}) in low dynamic
range has a certain amount of error, while that in high dynamic range
is correct and accurate. In our dataset, the rendering of shading and
specular residue images, as well as the estimation of specular-free
and specular highlight images, is carried out in high dynamic range.
And all generated images are converted to be of low dynamic range for
network training.

To overcome the above issues, we propose the Specular-free Refinement
stage (SR) to further refine the result from PSHR.
Figure~\ref{fig:framework}(b) illustrates the mechanism of SR. As
shown, the coarse specular-free image, along with the input, is fed
into an encoder-decoder network ($E_r$-$D_r$) to produce a refined
result. Compared to PSHR, SR is able to produce better results in
terms of detail preserving and natural appearances; see
Figure~\ref{fig:ablationstudy}(c). Furthermore, the histogram
comparisons of Figure~\ref{fig:ablationstudy}(e)(f) also validate the
performance improvement. As shown, compared to the coarse result, the
intensity distribution of the refined result is more consistent with
that of the input image over the non-highlight region.

\vspace{0.5em} \noindent \textbf{Loss Function}. The loss for
specular-free refinement $\mathcal{L}^{\text{SR}}$ is defined as
\begin{equation}
\mathcal{L}^{\text{SR}}=||D_2-\hat{D}||^2\,,
  \label{eq:HRloss}
\end{equation}
where $D_2$ and $\hat{D}$ are the refined specular-free image and
its ground truth, respectively.

\subsection{Tone Correction}
\label{subs:saturationcorrection}

Although the specular-free image from PSHR is further refined by SR,
its overall tone is sometimes noticeably different from the input, and
thus looks somewhat unreal; see Figure~\ref{fig:ablationstudy}(c). The
main reason is that the specular-free images in our training data are
of slightly lower brightness than the input images, due to the
inherent defect of software rendering; see Figure~\ref{fig:datashow}.
To overcome this issue, we propose the Tone Correction stage (TC) to
adjust the tone of the refined result to match that of the input as
closely as possible. Figure~\ref{fig:framework}(c) illustrates the
mechanism of TC. As shown, the refined result, along with the input
and specular residue images, is fed into an encoder-decoder network
($E_c$-$D_c$) to produce a tone-corrected result.
Figure~\ref{fig:ablationstudy} validate the effectiveness of TC in
terms of tone preservation. From it, we can see that the overall tone
of the tone-corrected result by TC is significantly closer to that of
the input than the results by PSHR and SR.

\begin{figure}[t]
  \centering
  \begin{tabular}{c}
    \makebox[0pt]{\includegraphics[width=0.99\columnwidth]{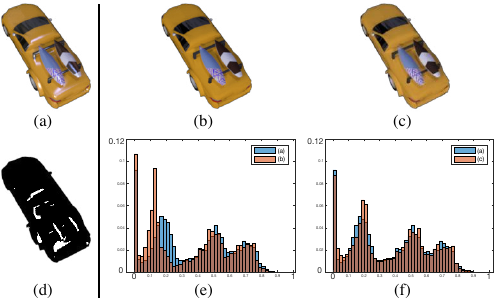}}
  \end{tabular}
  \caption{Tone correction for ground truth specular-free images
    in our dataset. (a) Input. (b) Ground truth specular-free
    image. (c) Tone correction version of (b). (d) Specular highlight
    mask of (a). (e) Histogram comparison between (a) and (b). (f)
    Histogram comparison between (a) and (c). Note that the specular
    highlight pixels are excluded using (d) for plotting histograms of
    (a)-(c).}
  \label{fig:tonecorrectionpipeline}
\end{figure}

The key idea of TC is to correct the tone of the ground truth
specular-free images in our dataset as new supervisions for network
training. Figure~\ref{fig:tonecorrectionpipeline} illustrates this
mechanism. Formally, given an input image $I$, and its ground truth
specular-free image $\hat{D}$ and specular residue image $\hat{R}$, we first use
Otsu's method on $\hat{R}$ to separate all pixels of $I$ into two types of
regions, specular highlight region $M_h$ and non-highlight region
$M_n$. Then, we find tone correction function $T$ that minimizes the
tone correction error $E$ between the specular-free and input images
over the non-highlight region:
\begin{equation}
E=|T(\hat{D})-I|_{\Omega_{M_n}}^2\,,
  \label{eq:saturationcorrectionformulation}
\end{equation}
where $\Omega_{M_n}$ denotes all pixels of $M_n$.
We formulate $T$ as the following linear transformation:
\begin{equation}
T=\mathbf{M}\ast
(p_h\ \ p_s\ \  p_v\ \ 1)'
\,,
  \label{eq:correctionfunction}
\end{equation}
where $p$ denotes a pixel in $\hat{D}$, whose intensity value in HSV
color space is $(p_h,p_s,p_v)$; $\textbf{M}$ is a $3\times 4$ matrix
which stores the parameters in the tone correction function; $\ast$
denotes matrix multiplication; and $(\cdot)'$ denotes matrix transpose.
The above operation in HSV instead of RGB benefits obtaining a robust
solution, because specular highlights mainly cause variations in the
saturation and value channels. We can solve the problem in
Eq.~(\ref{eq:saturationcorrectionformulation}) using the least-squares
method. Finally, we utilize $T$ to correct all pixels of $\hat{D}$ for
each training group in our dataset, and use them as new supervisions
for network training.

\vspace{0.5em} \noindent \textbf{Loss Function}. The loss for tone correction $\mathcal{L}^{\text{TC}}$ is
defined as
\begin{equation}
\mathcal{L}^{\text{TC}}=||D_3-\tilde{D}||^2\,,
  \label{eq:SCloss}
\end{equation}
where $D_3$ and $\tilde{D}$ are the tone-corrected specular-free
image and its ground truth, respectively.

\begin{figure*}[t]
  \centering
  \begin{tabular}{c}
    \makebox[0pt]{\includegraphics[width=0.99\textwidth]{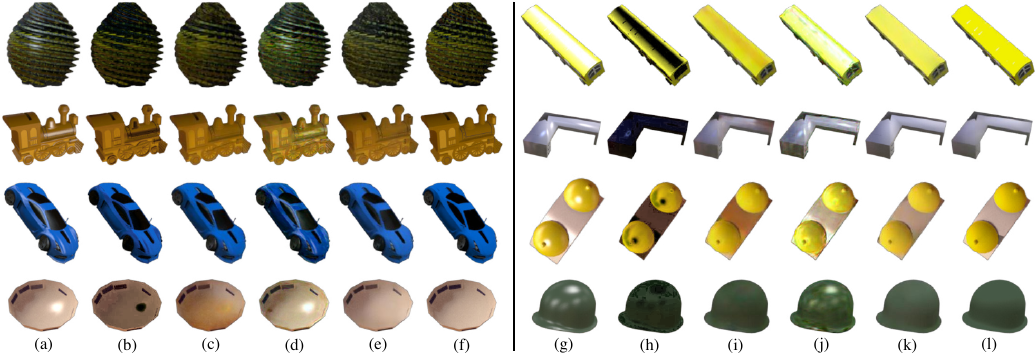}}
  \end{tabular}
  \caption{Visual comparison of our method against state-of-the-art
    methods on our synthetic testing images. (a)(g) Input. (b)(h) Yang
    \textit{et al.} \cite{yang-2015-effic-robus}. (c)(i) Fu \textit{et
      al.} \cite{fu-2021-multi-task}. (d)(j) Wu \textit{et al.}
    \cite{wu-2021-singl-image-a}. (e)(k) Ours. (f)(l) Ground truth.}
  \label{fig:resultsourdataset}
\end{figure*}

\subsection{Network Training}
\label{subs:networktraining}

The total loss $\mathcal{L}$ for the training of our whole network
includes $\mathcal{L}^{\text{PSHR}}$, $\mathcal{L}^{\text{SR}}$, and
$\mathcal{L}^{\text{TC}}$, written as
\begin{equation}
\mathcal{L}=\lambda_1 \mathcal{L}^{\text{PSHR}}+\lambda_2 \mathcal{L}^{\text{SR}}+\lambda_3\mathcal{L}^{\text{TC}}\,.
  \label{eq:totalloss}
\end{equation}
Here, $\lambda_1$, $\lambda_2$, and $\lambda_3$ are the weighting
balance parameters, which are experimentally set to $1$.

\subsection{Implementation Details}
\label{subs:implementation}

The four encoder-decoder networks in our three-stage framework have
the same architecture. We adopt the U-Net architecture
\cite{ronneberger-2015-u} as the default choice, known for its
conciseness and effectiveness. We implement our whole network in
PyTorch and train it for 60 epochs on a PC with NVIDIA GeForce GTX
3090Ti. The whole network is optimized using the Adam optimizer. The
initial learning rate is set to $1\times 10^{-4}$, divided by $10$
after every 10 epochs, and the batch size is set to $16$. Moreover, we
also adopt horizontal flip, and specular highlight attenuation and
boosting editing \cite{fu-2021-multi-task} for data augmentation.

\section{Experimental Results}
\label{sec:experiment}

\subsection{Datasets and Evaluation Metrics}
\label{subs:datasetmetrics}

We evaluate our network on three datasets, including our SSHR, SHIQ
\cite{fu-2021-multi-task}, and PSD \cite{wu-2021-singl-image-a}. We
adopt two commonly-used metrics (\textit{i.e.}, PSNR and SSIM) to
quantitatively evaluate the performance of our network, as in
\cite{fu-2021-multi-task, yang-2015-effic-robus}. In general, higher
PSNR and SSIM values indicate better results.

\subsection{Comparison with State-of-the-Art Methods}
\label{subs:comparionwithothers}

We compare our method against four traditional methods
\cite{tan-2005-separ-reflec, shen-2013-real-time,
  yang-2015-effic-robus, fu-2019-specul-highl} and two recent deep
learning-based methods \cite{fu-2021-multi-task,
  wu-2021-singl-image-a}. For fair comparison, we produce removal
results for four traditional methods using publicly available
implementation provided by the authors with optimal parameter setting.
Besides, if necessary, we re-train two deep learning-based methods,
and fine-tune their key parameters to produce better results as much
as possible. We note that our network fails to be trained on SHIQ and
PSD, since they do not include ground truth intrinsic images. To
train and evaluate our network on them, we modify the
first stage of our method to estimate the specular-free and specular
residue instead of the original albedo and shading.

\vspace{0.5em}\noindent\textbf{Quantitative Comparison}.
Tables~\ref{tab:quantiativeComparisonRemoval} reports the quantitative
comparison result on three datasets. As shown, overall, our method
achieves higher PSNR and SSIM values,
indicating that our method is superior to state-of-the-art methods. In
addition, four traditional methods \cite{tan-2005-separ-reflec,
  shen-2013-real-time, yang-2015-effic-robus, fu-2019-specul-highl}
achieve much higher PSNR and SSIM values on our synthetic dataset than
real SHIQ and PSD datasets. The reason is two-fold. First, these
methods are based on the dichromatic reflection model, and so does the
rendering of our synthetic dataset. As a result, they are capable of
addressing our synthetic images. Second, real specular highlights in
SHIQ and PSD may not be well characterized by an idealized image
formation model, while images in them are often with adverse factors
such as white material surfaces and heavy texture.

\vspace{0.5em}\noindent\textbf{Visual Comparison}.
Figure~\ref{fig:resultsourdataset} presents the visual comparison on
our testing images. We can see that for images with nearly white
material surfaces, traditional methods often produce unrealistic
results with severe visual artifacts such as color distortion (see the
$4^{th}$ row in (b)) and black color block (see the $1^{st}$ row in
(h)). Although the deep learning-based method
\cite{fu-2021-multi-task} is able to effectively remove specular
highlights and recover the missing details, it sometimes suffers from
color distortion artifacts (see the $4^{th}$ in (c)). Besides, the
deep learning method \cite{wu-2021-singl-image-a} fails to effectively
remove specular highlights (see the $1^{st}$ row in (d)), and may
produce unreasonable texture details (see the $2^{nd}$ row in (d)). In
comparison, our method is able to produce high-quality photo-realistic
removal results without noticeable visual artifacts caused by previous
methods. Due to space limit, the visual comparisons on SHIQ and
PSD are provided in our supplementary material.

\begin{table}[tbp]
  \small
  \centering
  \caption{Quantitative comparison of our method with state-of-the-art
    specular highlight removal methods on our SSHR,
    SHIQ \cite{fu-2021-multi-task}, and PSD
    \cite{wu-2021-singl-image-a}. The best results are marked in
    \textbf{bold}, while the second-best results are
    \underline{underlined}. Ours-A, Ours-B, and Ours-C denote our
    network without the specular-free refinement stage, the tone
    correction stage, and both these two stages, respectively.}
  \vspace{0.3cm}
  \label{tab:quantiativeComparisonRemoval}
  \setlength{\tabcolsep}{0.8mm}{
    \begin{tabular}{c|c|c|c|c|c|c}
      \hline
      Dataset &  \multicolumn{2}{c|}{SSHR} & \multicolumn{2}{c|}{SHIQ} &\multicolumn{2}{c}{PSD} \\
      \hline
      Metric &PSNR$\uparrow$ & SSIM$\uparrow$ &PSNR$\uparrow$ & SSIM$\uparrow$ & PSNR$\uparrow$ & SSIM$\uparrow$ \\
      \hline\hline
      Tan \cite{tan-2005-separ-reflec}  & 24.281             & 0.874             & 11.041             & 0.403             & 11.581             & 0.560             \\
      Shen \cite{shen-2013-real-time}   & 24.388             & 0.904             & 13.923             & 0.428             & 13.886             & 0.610             \\
      Yang \cite{yang-2015-effic-robus} & 23.243             & 0.894             & 14.310             & 0.502             & 12.866             & 0.611             \\
      Fu \cite{fu-2019-specul-highl}    & 23.270             & 0.881             & 15.746             & 0.723             & 14.400             & 0.665             \\
      Fu \cite{fu-2021-multi-task}      & 26.979             & 0.895             & \textbf{34.131}    & 0.860             & {21.516}           & 0.883             \\
      Wu \cite{wu-2021-singl-image-a}   & 25.731             & 0.894             & 23.420             & \underline{0.920} & \underline{21.801} & 0.880             \\
      \hline\hline
      Ours-A                            & 26.083             & 0.918             & 24.930             & 0.896             & {21.263}           & \underline{0.897} \\
      Ours-B                            & \textbf{28.903}    & \textbf{0.945}    & 22.019             & 0.843             & 18.932             & 0.830             \\
      Ours-C                            & 24.231             & 0.893             & 20.309             & 0.805             & 18.301             & 0.801             \\
      \hline\hline
      Ours                              & \underline{28.633} & \underline{0.940} & \underline{25.575} & \textbf{0.933} & \textbf{22.759}    & \textbf{0.903}    \\
      \hline
    \end{tabular}}
\end{table}

\begin{figure*}[t]
  \centering
  \begin{tabular}{c}
    \makebox[0pt]{\includegraphics[width=0.99\textwidth]{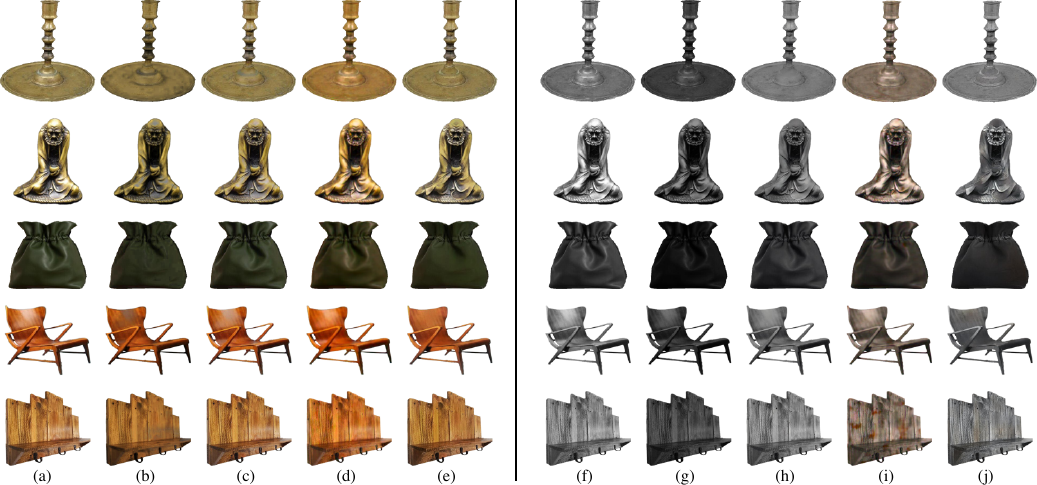}}
  \end{tabular}
  \vspace{0.005cm}
  \caption{Visual comparison of our method against state-of-the-art
    methods on real object-level images. (a)(f) Input and its
    grayscale version, respectively. (b)(g) Fu \textit{et al.}
    \cite{fu-2019-specul-highl}. (c)(h) Fu \textit{et al.}
    \cite{fu-2021-multi-task}. (d)(i) Wu \textit{et al.}
    \cite{wu-2021-singl-image-a}. (e)(j) Ours.}
  \label{fig:visualcomparisononobjectlevelimages}
\end{figure*}

\begin{figure*}[t]
  \centering
  \begin{tabular}{c}
    \includegraphics[width=0.92\textwidth]{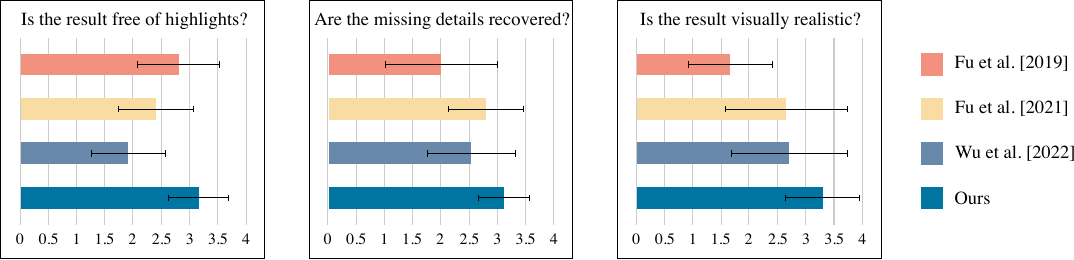}
  \end{tabular}
  \vspace{0.005cm}
  \caption{User study results on the three questions.}
  \label{fig:userstudy}
\end{figure*}

\vspace{0.5em}\noindent\textbf{User Study}. We further conducted a
user study to evaluate the robustness and generalization capability of
our method on real images. Here, three recent state-of-the-art
methods \cite{fu-2019-specul-highl,fu-2021-multi-task,
  wu-2021-singl-image-a} are compared. We first randomly downloaded
200 images from the Internet by searching the keywords ``chair'',
``statue'', ``storage bag'', and ``decoration''.
Figure~\ref{fig:visualcomparisononobjectlevelimages}(a) presents
several example images. Then, we produced specular-free images for all
downloaded images using our method and other compared methods, and
recruited 20 participants from a school campus for rating. Finally, we
asked the participants to score all results in a random order using a
1(worst)-to-4(best) scale (as done in \cite{wang-2019-under-photo,
  he-2019-progr-color}) on the three questions: (1) Is the result free
of highlights? (denoted as Q1); (2) Are the missing details recovered?
(denoted as Q2); and (3) Is the result visually realistic? (denoted as
Q3).

Figure \ref{fig:userstudy} summarizes the user study results, where
the average and standard deviation values of scores received by each
method are presented. As shown, our method achieves higher average
scores and lower standard deviations, indicating that our results are
more preferred by the participants with lower subjective bias.
Figure~\ref{fig:visualcomparisononobjectlevelimages} presents the
visual comparison on example images. As shown, our method is able to
effectively address real images and produce high-quality results with
natural appearances, although it is trained on the synthetic data.

\begin{figure*}[t]
  \centering
  \begin{tabular}{l}
    \makebox[0pt]{\includegraphics[width=0.99\textwidth]{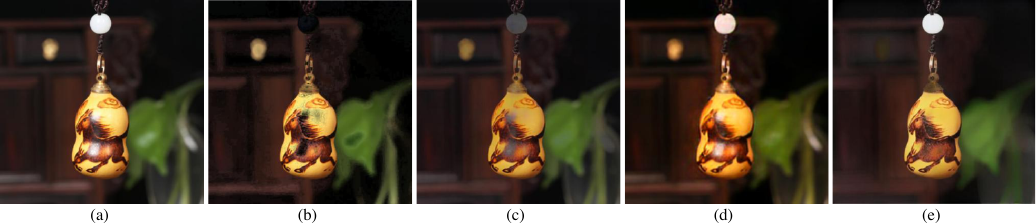}}
  \end{tabular}
  \vspace{0.005cm}
  \caption{Visual comparison of our method against
    state-of-the-art methods on a challenging image with multiple
    background objects. (a) Input. (b) Yang \textit{et al.}
    \cite{yang-2015-effic-robus}. (c) Fu \textit{et al.}
    \cite{fu-2021-multi-task}. (d) Wu \textit{et al.}
    \cite{wu-2021-singl-image-a}. (e) Ours.}
  \label{fig:visualcomparisonnaturalimages}
\end{figure*}

\subsection{Discussions}
\label{subs:discussions}

\noindent\textbf{Ablation Study}. Besides the visual comparison
results shown in Figure \ref{fig:ablationstudy}, we also
quantitatively validate the effectiveness of each stage of our method
(denoted as ``Ours'')
by constructing the following three variants:
\begin{itemize}
  \vspace{-0.1cm} 
\renewcommand\labelitemi{$\bullet$}
  \setlength{\itemsep}{0pt}
  \setlength{\parsep}{0pt}
  \setlength{\parskip}{0pt}
  \setlength{\parskip}{4pt}
\item Ours-A: ours without \textit{specular-free refinement}.
\item Ours-B: ours without \textit{tone correction}.
\item Ours-C: ours without both \textit{specular-free refinement}
  and \textit{tone correction} (\textit{i.e.}, only with \textit{physics-based
  specular highlight removal}).
  \vspace{-0.04cm} 
\end{itemize}

Table \ref{tab:quantiativeComparisonRemoval} reports the quantitative
results of our method and its variants on our SSHR, SHIQ, and PSD.
From the results, we can observe that the PSNR and SSIM scores of our
method and its three variants overall follow the relationship: Ours
$>$ Ours-A $>$ Ours-B $>$ Ours-C, except for a special case: Ours-B
$>$ Ours $>$ Ours-A $>$ Ours-C on our dataset. From it, we can draw
two conclusions. First, as the number of the used stages increases,
the performance of our method overall gets better and better,
illustrating the effectiveness of each stage of our method. Second,
the tone correction stage leads to a performance drop on our dataset,
due to the domain gap between our synthetic data and its tone
correction version. However, it further improves the performance on
SHIQ and PSD. This is because the resulting errors from the
differences between them and their tone correction versions can be
fully offset by the performance gain brought by further learning of
the network.

\vspace{0.5em}\noindent\textbf{Generalization to Grayscale Images}.
Figure~\ref{fig:visualcomparisononobjectlevelimages} presents the
visual comparison on color images (see the left column) and their
grayscale version (see the right column). As can be seen, the
traditional method \cite{fu-2019-specul-highl} suffers from leaking a
small amount of specular highlights into the specular-free images (see
the $1^{st}$ and $2^{nd}$ rows in (g)). For two deep learning-based
methods, the method \cite{fu-2021-multi-task} sometimes fails to
effectively remove specular highlights (see the $3^{rd}$ row in (h)).
The method \cite{wu-2021-singl-image-a} produces unsatisfactory or
even poor results with visual artifacts such as severe color distortion
and disharmonious color block. In comparison, our method trained on
our synthetic data is able to generalize well to real grayscale
images, which have almost the same performance as on color images.

\vspace{0.5em}\noindent\textbf{Generalization to Scene-Level Images}.
Figure \ref{fig:visualcomparisonnaturalimages} presents the visual
comparison on scene-level images. As can be seen, the traditional
method \cite{yang-2015-effic-robus} often mistakes white material
surfaces (see the circular jade in (b)) as specular highlights to be
removed, and undesirably produce black color block artifacts. For the two deep
learning-based methods, the method \cite{fu-2021-multi-task} fails to
effectively recover the missing color underneath specular highlights.
The method \cite{wu-2021-singl-image-a} often produces unsatisfactory
results with color distortion artifacts. In comparison, our method
produces good results with realistic color and clear texture
details. This illustrates that our method is able to generalize to
scene-level images with multiple background objects to a certain extent.

\begin{figure}[t]
  \centering
  \begin{tabular}{c}
    \makebox[0pt]{\includegraphics[width=0.99\columnwidth]{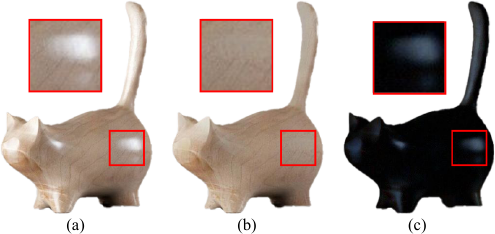}}
  \end{tabular}
  \vspace{0.005cm}
  \caption{A failure case of our method. (a) Input. (b)
    Specular-free image. (c) Specular residue image.}
  \label{fig:limitations}
\end{figure}

\vspace{0.5em}\noindent\textbf{Limitations}. Our method has two
limitations. First, our method, as well as previous methods, all
fail to recover missing texture details and color underneath
strong (\textit{i.e.}, high-intensity and large-area) specular highlights.
Figure~\ref{fig:limitations} presents an example. As can be seen, the
missing detailed patterns on the body of the wooden kitten underneath
strong specular highlights (see the red boxes) are less able to
be recovered very well. Second, although our method achieves good
results for object-level images, it may produce unsatisfactory
results, particularly for complex natural scenes often with achromatic
material surfaces, color lighting, noise, and so on.

\section{Conclusion}
\label{sec:conclusion}

We in this paper have proposed a three-stage method for object-level
specular highlight removal. Our key idea is to progressively eliminate
multiple types of visual artifacts to produce high-quality results
with natural appearances. In addition, we have presented a large-scale
synthetic dataset of object-level images to facilitate network training
and quantitative evaluation. In our dataset, each input specular
highlight image has corresponding ground truth albedo, shading,
specular residue, and specular-free images. We have conducted
extensive experiments to illustrate the superiority of our method over
previous methods in terms of quantitative comparison (\textit{i.e.},
higher PSNR and SSIM values), visual comparison, and a user study.

Our future work is to integrate features from inpainting
\cite{liu-2022-reduc} into our network to remove strong specular
highlights while restoring the missing texture details and color
underneath them. Another direction is to design more effective and
complex backbone networks such as diffusion models
\cite{ho-2020-denois-diffus} to further improve the performance of our
method.

\section*{Acknowledgments}
\label{sec:acknow}

This work was supported in part by the National Natural Science
Foundation of China under Grant 61972298, in part by the CAAI-Huawei
MindSpore Open Fund, and in part by The Hong Kong Polytechnic
University under Grants P0030419, P0043906, P0042740, and P0044520.

{\small
\bibliographystyle{ieee_fullname}
\bibliography{main_single.bib}
}

\end{document}